\documentclass[11pt]{article}

\usepackage{acl}

\usepackage{times}
\usepackage{latexsym}
\usepackage[T1]{fontenc}
\usepackage[utf8]{inputenc}
\usepackage{microtype}
\usepackage{inconsolata}
\usepackage{graphicx}
\usepackage{svg}
\usepackage{amsmath}
\usepackage{booktabs}
\usepackage{tabularx}
\usepackage{array}
\usepackage{enumitem}
\usepackage{xspace}
\usepackage{multirow}
\usepackage{fontawesome5}
\usepackage{xcolor}
\usepackage{url}

\definecolor{revisionblue}{RGB}{0,0,170}
\DeclareRobustCommand{\rev}[1]{#1}

\graphicspath{{workspace_state/experiments/multi_subject_paper_assets/figures/}{workspace_state/paper_screenshots/}{./}}

\newcommand{\BKT}{\textsc{BKT}\xspace}
\newcommand{\CI}{\textsc{CI}\xspace}
\newcommand{\DIME}{\textsc{DIME}\xspace}
\newcommand{\KReC}{\textsc{KReC}\xspace}

\newcommand{\Atlas}{Ontology Atlas\xspace}

\newcommand{\includevector}[2][]{%
  \IfFileExists{#2.pdf}{%
    \includegraphics[#1]{#2.pdf}%
  }{%
    \IfFileExists{#2.png}{%
      \includegraphics[#1]{#2.png}%
    }{%
      \includesvg[#1]{#2}%
    }%
  }%
}

\title{Evaluating Adaptive Personalization of Educational\\
Readings with Simulated Learners}

\author{
  \begin{tabular}{c}
    \\
    {\large\bfseries Ryan T. Woo\thanks{equal contribution.} \quad
    Anmol Rao\footnotemark[1] \quad
    Aryan Keluskar \quad
    Yinong Chen}\\[8pt]
    {\normalsize\normalfont School of Computing and Augmented Intelligence} \\
    {\normalsize\normalfont Arizona State University} \\
    {\normalsize\normalfont\texttt{\{rtwoo,arao75,akeluska,yinong.chen\}@asu.edu}}
  \end{tabular}
}

\begin{document}
\maketitle

\begin{abstract}
\rev{We present a framework for evaluating adaptive personalization of educational reading materials with theory-grounded simulated learners.} The system builds a learning-objective and knowledge-component ontology from \rev{Wikibooks} open textbooks, curates it in a browser-based Ontology Atlas, labels textbook chunks with ontology entities, and generates aligned reading--assessment pairs. Simulated readers learn from passages through a Construction--Integration-inspired memory model with \DIME-style reader factors, \KReC-style misconception revision, and an open New Dale--Chall readability signal. Answers are produced by score-based option selection over the learner's explicit memory state, while \BKT drives adaptation. \rev{Under this explicit simulator parameterization and three sampled subject ontologies,} adaptive reading significantly improved outcomes in computer science, yielded smaller positive but inconclusive gains in inorganic chemistry, and was neutral to slightly negative in general biology.
\end{abstract}

\section{Introduction}

With more students entering higher education, instructors increasingly face a familiar mismatch: students are expected to learn the same material at the same pace even though their understanding of particular concepts can vary widely. In an introductory course, one student may already understand a subset of the target concepts, while another may still hold robust misconceptions about those same concepts. When both students are given identical readings, examples, and explanations, one may be under-challenged while the other is overwhelmed. Addressing this in a real classroom typically requires instructors to rewrite materials, manually identify which concepts students struggle with, and provide individualized support. Those interventions are difficult to scale.

Intelligent tutoring systems (ITSs) aim to address this problem by adapting instruction to the learner. Recent reviews suggest that ITSs often improve learning and performance, but that their impact depends strongly on the instructional setting, the adaptation mechanism, and the way systems are evaluated \citep{huang2025effects, letourneau2025systematic}. In many ITS settings, however, the core intervention is an item sequence, a hint policy, or post-hoc feedback. Our setting is different: the primary intervention is \emph{reading material}. Questions are still important, but mainly because they reveal evidence about what the learner knows after reading and provide observations for mastery tracking.

To evaluate whether adaptive reading policies help, we need simulated learners that can (i) learn from text, (ii) differ in reading-related characteristics, (iii) preserve misconceptions rather than merely lacking knowledge, and (iv) later answer multiple-choice questions from what they remember rather than from unrestricted access to the source passage. Simulated learners have long been proposed as tools for teachers, for learners, and for instructional designers conducting formative evaluation \citep{vanlehn1994applications}. More recently, a systematic review across AIED-related fields found uses spanning teacher training, adaptive algorithm development, virtual collaborators, and testing of learning environments, while also concluding that realism and validity are often under-tested \citep{kaeser2024simulated}. Off-the-shelf LLM prompting is a poor fit for that requirement because powerful LLMs often answer beyond the intended ability of a partially knowledgeable learner. \citet{yuan2026towards} characterize this problem as a \emph{competence paradox} and argue for constrained student simulation under an explicit epistemic state specification.

We therefore present a theory-grounded simulated reader for adaptive reading personalization. The system builds a learning-objective (LO) and knowledge-component (KC) ontology from open textbooks, tracks KC mastery with \BKT \citep{corbett1994knowledge}, generates ontology-driven reading and assessment variants, and simulates reading via a comprehension model grounded in the Construction--Integration (\CI) theory of discourse comprehension \citep{kintsch1988role}. Learner heterogeneity is parameterized with \DIME-style reader factors \citep{cromley2007testing, cromley2010reading}, misconceptions are revised through the Knowledge Revision Components framework (\KReC) \citep{kendeou2014krec, kendeou2024theory}, and later answer selection is performed through score-based option selection over an explicit epistemic boundary.

\rev{The experiments focus on offline simulation so that the instructional policy can be studied before classroom deployment. This setup compares adaptive and non-adaptive reading policies under matched simulated cohorts while keeping the content pipeline, question format, and mastery tracker fixed.}

Our main contributions are as follows:
\begin{itemize}[leftmargin=1.2em]
    \item We formulate adaptive reading personalization as a closed-loop problem in which reading is the primary intervention and assessment is the observation channel for KC-level mastery tracking.
    \item We present an ontology-grounded content pipeline that links open textbook chunks, reading variants, and assessment items to shared LOs and KCs.
    \item We develop a simulated reader that combines \CI, \DIME, \KReC, and an open New Dale--Chall readability signal to model what a learner comprehends, revises, retains, and later retrieves.
    \item We report multi-subject experimental results showing that adaptive reading yields domain-dependent gains: statistically reliable improvements in the sampled computer-science setting, positive but inconclusive gains in inorganic chemistry, and neutral to slightly negative outcomes in general biology.
\end{itemize}

\section{Related Work}

\subsection{Adaptive tutoring and knowledge tracing}

Knowledge tracing estimates latent mastery from observed responses, and \BKT remains attractive because it is interpretable at the level of individual KCs \citep{corbett1994knowledge}. Although deep models can improve prediction, interpretability remains important for deployment and diagnosis \citep{ding2021interpretability}. We therefore use \BKT so that the state driving adaptation remains understandable.

Most ITS personalization focuses on problem selection, hints, or feedback. By contrast, we center on explanatory reading material. Because reading can change knowledge without explicit responses, we separate the learner's hidden \emph{true} knowledge from the tutor's \BKT-based \emph{estimate}, and only the latter drives adaptation.

\subsection{Simulated learners in educational technology}

Simulated learners have long been used in educational technology. \citet{vanlehn1994applications} categorize their use for teachers, learners, and instructional designers; our setting aligns most closely with formative evaluation of instructional policies before deployment.

Recent work highlights broader uses, including teacher training, hypothesis generation, adaptive algorithm development, virtual collaborators, and environment testing, while also noting that many simulators remain under-validated for their intended decisions \citep{kaeser2024simulated}. This motivates our emphasis on explicit learner state, inspectable memory, and theory-grounded reading dynamics.

\subsection{Reading comprehension and computational reader models}

Our model is grounded in the Construction--Integration framework \citep{kintsch1988role}, where readers form a propositional \emph{textbase} and a richer \emph{situation model} integrating prior knowledge. This supports modeling partial retention, revision, and distortion of information.

We extend \CI with \DIME, which captures reader differences in background knowledge, vocabulary, and inference \citep{cromley2007testing, cromley2010reading}, and \KReC, which models revision of incorrect prior knowledge \citep{kendeou2014krec, kendeou2024theory}. This matters because many systematic wrong answers reflect stable misconceptions rather than random noise \citep{vandenbroek2008coactivation, gierl2017developing}.

Prior work has implemented \CI-like automated readers \citep{lemaire2006computational}; our contribution is to embed such a reader in a KC-level adaptive instructional loop.

\subsection{Epistemic boundaries and answer modeling}

Recent work argues that valid student simulation should expose an explicit epistemic state and bounded competence \citep{yuan2026towards}. We adopt that principle, but do not use free-form generation for answer choice. Instead, answers are selected by scoring options from the learner's memory state, including mastery, misconceptions, retrieved traces, and item difficulty. This design is efficient, auditable, and consistent with constrained learner behavior. \rev{It also favors inspectable state and reproducible repeated simulations over unconstrained generative answers.}

\subsection{Readability and reader--text matching}

Reader--text fit is important but should remain simple to operationalize. We use the New Dale--Chall readability formula \citep{chall1995readability}, computed via \texttt{textstat} \citep{textstat2026}, as a reproducible measure of text difficulty combined with learner parameters. \rev{The formula is treated only as a lightweight signal; its tradeoffs are discussed in the limitations.}

\section{System Overview}

\begin{figure*}[t]
    \centering
    \includegraphics[width=0.96\textwidth]{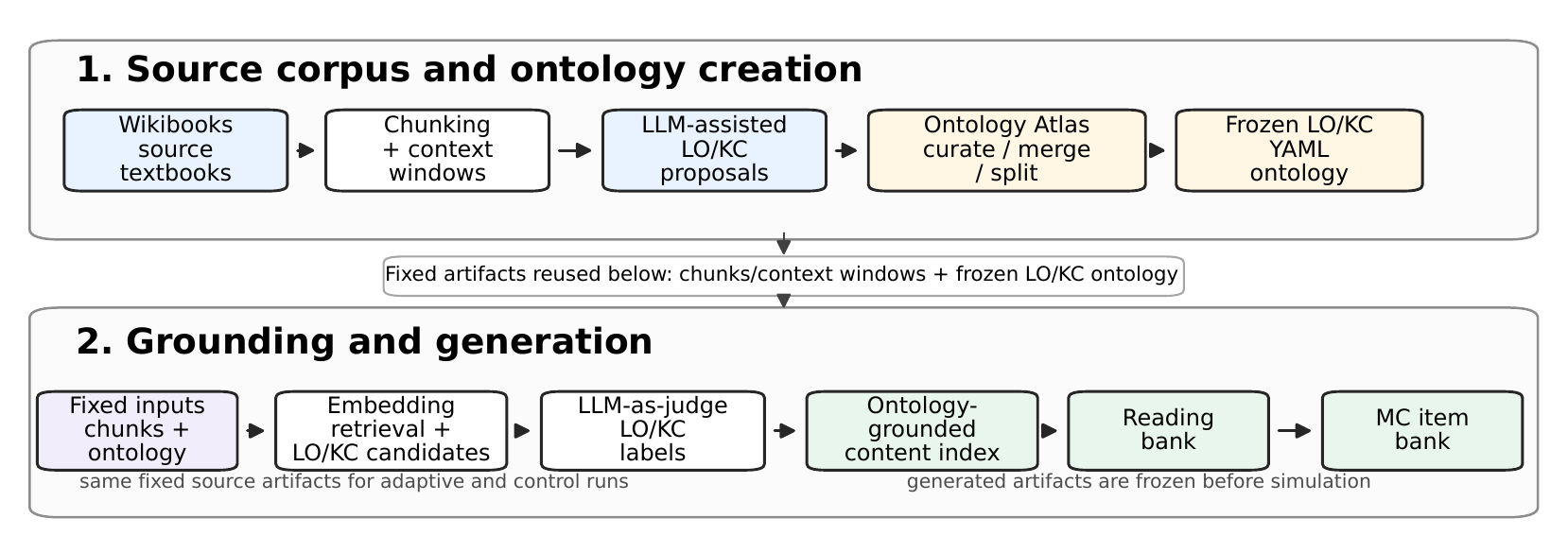}
    \caption{\rev{Fixed source-corpus, ontology, grounding, and generation pipeline used by both conditions.}}
    \label{fig:content-pipeline}
\end{figure*}

\begin{figure}[t]
    \centering
    \includegraphics[width=0.96\linewidth]{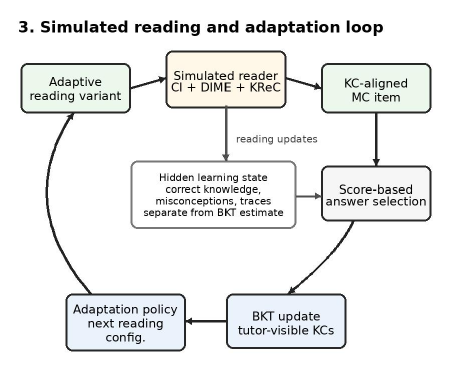}
    \caption{\rev{Closed simulated reading and adaptation loop.}}
    \label{fig:simulation-loop}
\end{figure}

The overall system converts open educational content into a grounded personalization loop. The pipeline can be summarized in five stages: (1) ontology construction and authoring, (2) chunk labeling and retrieval indexing, (3) ontology-driven reading and assessment generation, (4) theory-grounded simulated reading and score-based answer selection, and (5) \BKT-based adaptation. \rev{\autoref{fig:content-pipeline} separates the fixed source-corpus, ontology, grounding, and generation stages from the online simulation. \autoref{fig:simulation-loop} shows the simulated reading and adaptation loop as a closed cycle.}

\rev{The experiments use Wikibooks source textbooks: \emph{Foundations of Computer Science}, \emph{General Biology}, and \emph{Introduction to Inorganic Chemistry} \citep{wikibooks_fcs,wikibooks_general_biology,wikibooks_inorganic_chemistry}. These books serve as course-specific corpora: the ontology, retrieval index, generated readings, and assessments are grounded in the same instructional source material.}

\subsection{Ontology Atlas and ontology construction}

We first build an ontology of learning objectives and knowledge components from open textbooks. Each LO maps to one or more KCs. The ontology is created with an LLM-assisted workflow that proposes candidate LOs and KCs from source material and then consolidates them into a course-specific YAML schema. We use the ontology as a common representational layer for retrieval, generation, mastery tracking, and simulation.

\rev{The ontology construction process is semi-automated rather than fully automatic. LLM-generated LO and KC candidates are treated as proposals that can be accepted, merged, split, renamed, or deleted in Atlas before experiment artifacts are generated. Merge decisions are appropriate when two KCs represent the same assessable concept at the same grain size; split decisions are appropriate when a proposed KC combines multiple independently assessable skills or concepts. In the reported workflow, ontology files are fixed within each subject-level adaptive-control comparison so that the two policies use the same ontology.}

To make this layer inspectable, we built \Atlas, a browser-based ontology workspace backed directly by the YAML files used elsewhere in the system. Atlas exposes a coverage inventory showing chapter, LO, and KC counts; a searchable navigator over IDs, statements, labels, and descriptions; a D3 relationship graph for the chapter $\rightarrow$ LO $\rightarrow$ KC hierarchy; and a details panel for editing entries, importing/exporting YAML, validating fields, and saving back to disk. Atlas is therefore not just a visualization layer; it is the authoring interface for the ontology that the generation and experiment services consume.

\begin{figure*}[t]
    \centering
    \includegraphics[width=0.90\textwidth]{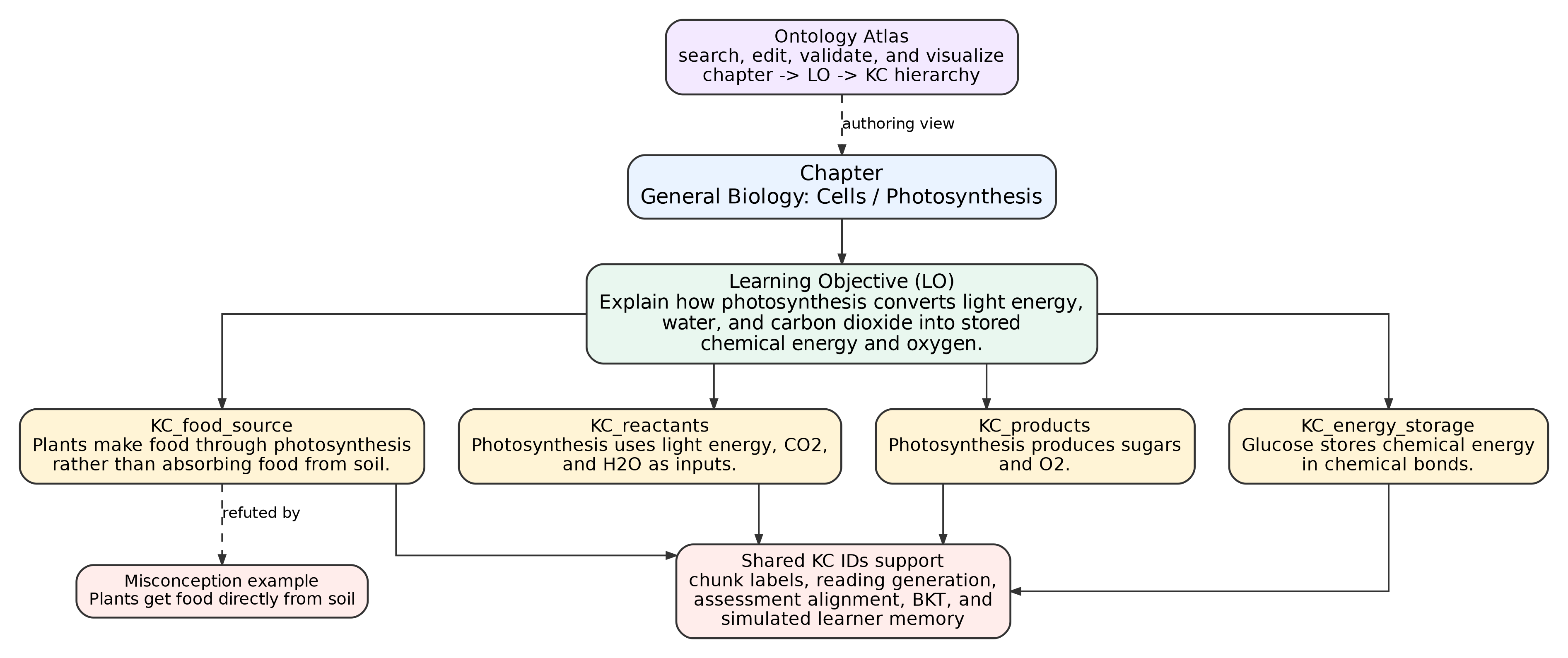}
    \caption{\rev{Example chapter $\rightarrow$ LO $\rightarrow$ KC relation visualized by \Atlas for photosynthesis.}}
    \label{fig:photosynthesis-ontology}
\end{figure*}

In the current codebase, \Atlas surfaces multiple full-course ontologies, including the three used in the experiments reported here: foundations of computer science (16 chapters, 53 LOs, 131 KCs), general biology (20 chapters, 60 LOs, 172 KCs), and introductory inorganic chemistry (12 chapters, 57 LOs, 177 KCs). This coverage view was useful when selecting sampled LOs for the experiments and when checking whether KC granularity was comparable across subjects.

\subsection{Chunk labeling and retrieval indexing}

The textbook corpus is segmented into fine-grained chunks and coarser context chunks. Fine-grained chunks serve as the primary evidence units for labeling, while coarser chunks provide broader context during downstream generation. Each fine-grained chunk is labeled with one or more LOs and KCs using a three-stage LLM-as-judge pipeline:
\begin{enumerate}[leftmargin=1.2em]
    \item encode the chunk as a semantic embedding,
    \item retrieve candidate LO and KC labels with nearest-neighbor search in a vector database, and
    \item prompt an LLM to adjudicate among the candidates and return the final label set.
\end{enumerate}

The result is a grounded index in which every retained content unit is tied to the ontology and remains linked to its surrounding context. Because Atlas edits the same YAML that the labeling and generation services read, ontology revisions immediately propagate to retrieval and generation.

\subsection{Ontology-driven reading and assessment generation}

Using the labeled corpus, we generate paired readings and assessments that comprehensively cover one LO at a time. Each generated reading is grounded in retrieved fine-grained chunks and optional coarser context chunks; generation parameters control explanation depth, example density, misconception refutation, and target difficulty. Each assessment is a fresh isomorphic multiple-choice variant aligned to the same LO and KCs rather than an exact reuse of the same item surface form.

This choice is important for two reasons. First, exact question reuse would make policy evaluation more sensitive to item wording than to learner state. Second, our simulated learners do not learn from assessment interactions; only reading updates their hidden semantic state. Using fresh isomorphic variants therefore preserves comparability across iterations without allowing assessment exposure to become an unintended instructional channel.

\rev{For preprocessing, semantic embeddings used \texttt{qwen3-embedding-8b}; LLM-assisted ontology proposal, chunk-label adjudication, reading generation, and assessment generation used \texttt{qwen3-30b-a3b-instruct-2507}. Once generated, the ontology, labels, readings, and assessments are treated as fixed artifacts for the adaptive-control comparison.}

\section{Methodology}

\subsection{Separating hidden learning from observed mastery}

The simulated learner maintains two distinct states. The first is a hidden \emph{true learning state} that governs what the learner actually knows, misunderstands, and remembers. The second is the tutor-visible \BKT estimate that is updated only from observed answers. This separation prevents the simulation from becoming self-fulfilling: the personalization policy acts on what the tutor \emph{infers}, not on hidden ground truth.

For learner $i$ and KC $k$, we represent hidden state as
\begin{equation}
    z_i = \{m^{*}_{i,k},\, w_{i,k,1:M},\, r_i,\, s_i,\, e_i\},
\end{equation}
where $m^{*}_{i,k}$ is the strength of correct knowledge for KC $k$, $w_{i,k,1:M}$ are strengths of KC-specific misconceptions, $r_i$ is a reader profile, $s_i$ is a revision-and-response profile, and $e_i$ stores recent episodic traces from readings. The tutor only observes a separate KC mastery estimate $\hat{m}_{i,k}$ maintained by \BKT.

\begin{table*}[t]
\centering
\small
\begin{tabularx}{\textwidth}{>{\raggedright\arraybackslash}p{2.3cm} >{\raggedright\arraybackslash}X >{\raggedright\arraybackslash}p{4.2cm}}
\toprule
\textbf{State component} & \textbf{Meaning} & \textbf{Role in the simulation} \\
\midrule
$ m^{*}_{i,k} $ & Hidden strength of correct knowledge for KC $k$ & Governs durable support for correct answers and future learning from related readings \\
$ w_{i,k,m} $ & Hidden strength of misconception $m$ for KC $k$ & Attracts misconception-consistent distractors and resists revision when refutation is weak \\
$ r_i $ & Reader profile (background knowledge, vocabulary, inferencing, strategy use, readability ability) & Determines how well teaching events are encoded from text \\
$ s_i $ & Revision and response profile (skepticism, revision willingness, detail preference, noise) & Controls how new text changes prior beliefs and how answers are emitted \\
$ e_i $ & Episodic traces of recently read teaching events & Supports short-term retrieval even when durable KC mastery is still weak \\
$ \hat{m}_{i,k} $ & Tutor-visible \BKT mastery estimate & Drives reading adaptation but does not directly change hidden learning \\
\bottomrule
\end{tabularx}
\caption{Learner-state variables.}
\label{tab:learner-state}
\end{table*}

\subsection{Reader profiles and readability matching}

We do not use vague ``learning styles.'' Instead, learner heterogeneity is parameterized through theory-grounded reader profiles. Following \DIME, each learner has parameters for background knowledge, vocabulary, inferencing, and strategy use \citep{cromley2007testing, cromley2010reading}. Because the cohort is intended to approximate college freshmen who have completed secondary school, basic decoding is assumed to be near ceiling; the meaningful variability is in comprehension-relevant factors and in misconception prevalence.

We combine these learner-side parameters with a reader--text fit term based on the New Dale--Chall score. \rev{The score is used as a compact, reproducible difficulty signal alongside the richer learner-side profile.}

For passage $j$, let $d_j$ be the readability score computed with \texttt{textstat}. Let $a_i$ be a scalar readability-ability parameter for learner $i$. We define
\begin{equation}
    \text{match}_{ij} = 1 - \frac{|a_i - d_j|}{6},
\end{equation}
clipped to $[-1,1]$, so higher values indicate a better match between learner and text. This term does not replace the richer reader profile; it provides a compact difficulty signal that can be integrated into the comprehension update.

\subsection{From passages to proposition-like teaching events}

We model reading at the level of comprehension and memory, not time. Each generated passage is segmented into short teaching events, which serve as proposition-like units for the simulation. The current implementation uses a lean operational schema rather than full semantic parsing: each event stores its surface text, linked KC IDs, a clarity score, a refutation-strength score, and a refresh flag.

These fields have direct roles in the simulation. \emph{KC links} determine which latent mastery variables and misconception weights the event can update. \emph{Clarity} approximates how easy the event is to encode; in the current implementation it is estimated from event length, so shorter events are treated as easier to encode than longer ones. \emph{Refutation strength} approximates how explicitly the event confronts a likely misconception; in the current implementation it is triggered by lexical cues such as \texttt{misconception}, \texttt{incorrectly}, \texttt{rather than}, \texttt{instead of}, or \texttt{do not}, and it scales misconception reduction during learning. \emph{Refresh} marks lower-gain reinforcement events when review KCs are included in a passage; operationally, refresh is a lightweight heuristic rather than a discourse parser.

This representation keeps learner memory close to the \CI notion of a textbase while remaining simple enough to generate, score, and inspect at scale. Encoded events create episodic traces that preserve the event text, KC links, clarity, refutation strength, readability score, and later distortion, making subsequent answer behavior inspectable. When provenance is needed, it is retained at the reading-asset level through source chunk identifiers rather than per-event source spans.

\rev{For example, the sentence pair ``Plants do not get food from soil. They make food through photosynthesis.'' is stored as a proposition-like teaching event linked to a photosynthesis KC:}
\begin{quote}
\footnotesize
\begin{verbatim}
{
proposition_id: "prop_001",
text: "Plants do not get food from soil.
They make food through photosynthesis.",
kc_ids: {"KC_photosynthesis_food_source"},
clarity: 0.92,
refutation_strength: 1.00,
is_refresh: false
}
\end{verbatim}
\end{quote}
\rev{The record is operational rather than a full semantic proposition.}

\subsection{Comprehension as textbase construction and situation-model integration}

The reading update proceeds in two stages. First, the learner encodes teaching events into a textbase-like episodic memory. Second, the learner integrates those events with prior knowledge to update correct knowledge and misconception weights. Operationally, the encoding strength of event $p$ for learner $i$ reading passage $j$ is modeled as
\begin{equation}
\label{eq:encoding}
\begin{split}
\text{enc}_{i,p} = \sigma\bigl(
&\alpha + 1.1\,\text{clarity}_p + 0.55\,\text{match}_{ij} \\
&{}+ \beta^\top r_i + 0.3(1-\bar{m}_{i,p}) \\
&{}+ 0.35\,\text{attention}_i
\bigr).
\end{split}
\end{equation}
Here, $\bar{m}_{i,p}$ is mean prior mastery over the event's target KCs and $\beta^\top r_i$ collects vocabulary, inference, strategy, and background-knowledge effects. If an event is encoded, it creates an episodic trace whose later activation depends on recency, rehearsal, lexical overlap with the question, and distortion noise.

Once encoded, an event updates KC state by increasing correct knowledge and reducing misconception weight. In the implementation, correct-knowledge gain is stronger for clearer events, better reader--text matches, and learners with higher revision willingness and detail preference. Misconception reduction is scaled by the event's refutation strength, consistent with \KReC's emphasis on how prior incorrect knowledge is confronted during reading \citep{kendeou2014krec, kendeou2024theory}. Refresh events receive a smaller gain multiplier than new target events.

\rev{The coefficients in \autoref{eq:encoding} and \autoref{eq:utility} were selected during simulator development and then held fixed across subject comparisons. They encode qualitative assumptions rather than fitted human-data parameters: clarity, match, relevant prior knowledge, attention, and strategy use increase encoding; misconception strength and item difficulty reduce correct responding; and retrieved traces provide partial support.}

\subsection{Score-based answer option selection from learner memory}

At assessment time, the learner does \emph{not} receive the original passage. Instead, the simulator retrieves relevant traces from memory and computes a score for each item from hidden mastery, misconception weight, retrieved-trace support, and response-style variables. The guiding principle is still epistemic boundedness in the sense of \citet{yuan2026towards}: the learner answers only from what it has encoded and retained. However, the current implementation uses score-based option selection rather than free-form generative reasoning.

For item $q$, let $\bar{m}_{iq}$ be mean correct knowledge over the target KCs, $\bar{w}_{iq}$ the mean misconception weight, and $\tau_{iq}$ the mean activation of retrieved traces. The item's response utility is
\begin{equation}
\label{eq:utility}
\begin{split}
u_{iq} ={}&
-b_q + 2.2\,\bar{m}_{iq} + 0.38\,\tau_{iq} \\
&{}- 1.25\,\bar{w}_{iq} + 0.35\,\text{attention}_i \\
&{}- 0.25\,\text{guess}_i,
\end{split}
\end{equation}
where $b_q$ is a difficulty offset derived from the item's difficulty band. The probability of a correct answer is then
\begin{equation}
P(y_{iq}=1)=\sigma(u_{iq}),
\end{equation}
clipped to a small interior interval to avoid degenerate probabilities.

If the learner answers incorrectly, the selected distractor is \emph{not} sampled uniformly. Instead, distractors are scored by lexical overlap between the retrieved trace text and the distractor text/rationale, plus learner-specific guessing bias. This makes misconception-consistent wrong options more likely than random errors when the learner's memory traces support them. The simulator also records an explicit epistemic-state summary for each response, including target KCs, correct-knowledge support, misconception support, and the IDs of the retrieved traces used during scoring.

\section{Adaptive Personalization with Bayesian Knowledge Tracing}

The tutor maintains a separate \BKT model for each KC. After each question response, the tutor updates the KC-specific mastery estimate $\hat{m}_{i,k}$ using standard \BKT parameters for initial mastery, learning, guess, and slip \citep{corbett1994knowledge}. These estimates drive reading adaptation but do not directly modify hidden learner state.

\rev{We use \BKT as the tutor-visible observation model, not as the cognitive model of reading. Passages update hidden semantic state; quiz responses provide noisy evidence for KC mastery estimates. \BKT is useful here because it gives the adaptation policy an interpretable KC-level state, so we report both hidden-state summaries and tutor-side \BKT gain.}

\rev{The adaptation policy maps low estimated mastery to more supportive readings. In the adaptive condition, low $\hat{m}_{i,k}$ can increase explanatory depth, example density, explicit misconception refutation, repeated weak-KC coverage, and reader--text matching. Higher $\hat{m}_{i,k}$ yields leaner variants. The control condition uses the same ontology, source corpus, and assessment machinery, but holds the reading configuration fixed rather than conditioning it on $\hat{m}_{i,k}$.}

\section{Experimental Setup}

We ran fixed-iteration experiments in three subject ontologies: foundations of computer science, general biology, and introductory inorganic chemistry. \autoref{tab:atlas-scale} summarizes the ontology sizes surfaced in \Atlas for those three subjects.

\begin{table}[t]
\centering
\small
\begin{tabular}{lrrr}
\toprule
\textbf{Subject} & \textbf{Ch.} & \textbf{LOs} & \textbf{KCs} \\
\midrule
Computer Science & 16 & 53 & 131 \\
General Biology & 20 & 60 & 172 \\
Inorganic Chemistry & 12 & 57 & 177 \\
\bottomrule
\end{tabular}
\caption{Ontology sizes shown in the \Atlas coverage view for the three ontologies used in the experiments.}
\label{tab:atlas-scale}
\end{table}

For each subject, we sampled four LOs across two chapters, created 50 matched simulated learners per condition, ran three reading--assessment cycles per LO, and generated three KC-aligned multiple-choice items per cycle. This yields 1{,}800 item-level responses per condition per subject.

\rev{Adaptive and control runs were matched at initialization: each control run reused the same learner seeds and initial hidden states as its adaptive counterpart. Assessment delivery context was set to \texttt{summative}, so quiz answers updated tutor-visible \BKT state but did not directly change hidden semantic knowledge. We report observed accuracy, tutor-side \BKT gain, hidden mastery gain, and misconception reduction. Because learner-level paired trajectories are preserved only for accuracy and \BKT gain, inferential tests are limited to those outcomes.}

\section{Results}

\subsection{Subject-level outcomes}

\begin{figure}[t]
    \centering
    \includegraphics[width=1\linewidth]{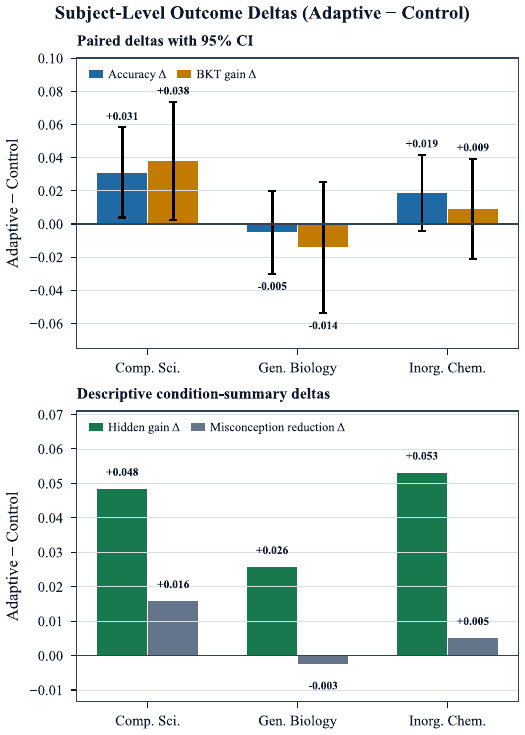}
    \caption{Subject-level adaptive-minus-control deltas.}
    \label{fig:subject-deltas}
\end{figure}

\rev{\autoref{fig:subject-deltas} and \autoref{tab:cycle-deltas} summarize adaptive-minus-control deltas. Computer science shows the clearest benefit: accuracy improved from 0.834 to 0.865 ($\Delta=+0.031$, 95\% CI [$+0.004$, $+0.058$], $p=0.026$), and learner-level \BKT gain improved from 0.416 to 0.454 ($\Delta=+0.038$, 95\% CI [$+0.002$, $+0.074$], $p=0.037$). Chemistry was positive but inconclusive (accuracy $\Delta=+0.019$, $p=0.104$; \BKT gain $\Delta=+0.009$, $p=0.540$). Biology was neutral to slightly negative (accuracy $\Delta=-0.005$, $p=0.691$; \BKT gain $\Delta=-0.014$, $p=0.468$). Descriptive hidden-state endpoints in \autoref{tab:mastery-misconception-change} favored adaptation for hidden mastery in all three subjects and for misconception reduction in computer science and chemistry.}

\begin{table}[!htbp]
\centering
\resizebox{\linewidth}{!}{%
\begin{tabular}{lccccccc}
\toprule
\multirow{2}{*}{Subject} & \multirow{2}{*}{Cycle} 
& \multicolumn{3}{c}{Observed Accuracy} 
& \multicolumn{3}{c}{Tutor-side BKT gain} \\
\cmidrule(lr){3-5} \cmidrule(lr){6-8}
& & Adaptive & Control & $\Delta$ 
& Adaptive & Control & $\Delta$ \\
\midrule

\multirow{3}{*}{Computer Science}
& 1 & 0.802 & 0.760 & +0.042 & 0.180 & 0.157 & +0.024 \\
& 2 & 0.890 & 0.853 & +0.038 & 0.168 & 0.146 & +0.023 \\
& 3 & 0.904 & 0.890 & +0.014 & 0.106 & 0.114 & -0.008 \\
\midrule

\multirow{3}{*}{General Biology}
& 1 & 0.794 & 0.795 & -0.001 & 0.174 & 0.167 & +0.007 \\
& 2 & 0.864 & 0.849 & +0.015 & 0.122 & 0.111 & +0.011 \\
& 3 & 0.877 & 0.906 & -0.029 & 0.073 & 0.106 & -0.033 \\

\midrule

\multirow{3}{*}{Inorganic Chemistry}
& 1 & 0.807 & 0.797 & +0.010 & 0.190 & 0.190 & +0.000 \\
& 2 & 0.868 & 0.828 & +0.040 & 0.146 & 0.136 & +0.010 \\
& 3 & 0.885 & 0.878 & +0.007 & 0.092 & 0.092 & +0.000 \\

\bottomrule
\end{tabular}}
\caption{Subject-level adaptive versus control outcomes.}
\label{tab:cycle-deltas}
\end{table}

\subsection{Cycle-level trajectories}

\begin{figure}[t]
    \centering
    \includegraphics[width=1\linewidth]{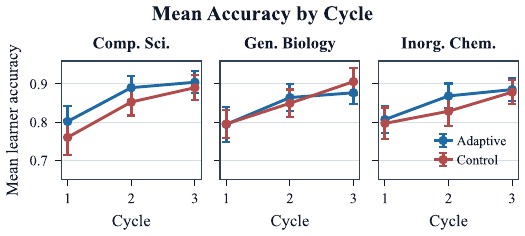}
    \caption{Mean learner-level quiz accuracy by fixed iteration cycle.}
    \label{fig:cycle-accuracy}
\end{figure}

\rev{\autoref{fig:cycle-accuracy} shows that adaptive reading led control in every computer-science and chemistry cycle, with the largest chemistry margin in cycle 2. Biology showed a crossover pattern: nearly tied in cycle 1, adaptive ahead in cycle 2, and control ahead in cycle 3.}

\begin{table}[!htbp]
\centering
\resizebox{\linewidth}{!}{%
\begin{tabular}{lcccccc}
\toprule
\multirow{2}{*}{Subject} & \multicolumn{3}{c}{Hidden mastery gain} 
& \multicolumn{3}{c}{Misconception reduction} \\
\cmidrule(lr){2-4} \cmidrule(lr){5-7}
& Adapt. & Ctrl. & $\Delta$ 
& Adapt. & Ctrl. & $\Delta$ \\
\midrule
Computer Science      & 0.521 & 0.473 & +0.049 & 0.072 & 0.056 & +0.016 \\
General Biology       & 0.388 & 0.362 & +0.026 & 0.045 & 0.048 & -0.003 \\
Inorganic Chemistry   & 0.509 & 0.456 & +0.053 & 0.061 & 0.056 & +0.005 \\
\bottomrule
\end{tabular}%
}
\caption{Descriptive hidden-state outcomes from saved experiment summaries.}
\label{tab:mastery-misconception-change}
\end{table}

\section{Discussion}

\rev{The results suggest domain-dependent effects of the adaptive policy. Computer science provides the most consistent evidence of benefit, chemistry shows smaller positive but inconclusive gains, and biology shows a mixed pattern in which hidden mastery improved descriptively without improving observed accuracy or tutor-side \BKT gain.}

This divergence between hidden and observed outcomes is informative. In the simulator, answers depend on hidden KC knowledge, episodic traces, misconception weight, item difficulty, attention stability, and distractor overlap. A reading can therefore improve hidden knowledge while producing little observed gain if retrieval is brittle, items are noisy, or distractors remain competitive.

\rev{The results also illustrate the role of \Atlas in auditing ontology granularity and coverage. Adaptive benefit varied by LO and subject, so ontology quality and KC granularity are not merely upstream bookkeeping issues. If an LO is too broad, if its KCs are diffuse, or if retrieved evidence is weakly aligned with target KCs, both personalization and evaluation become harder to interpret.}

\section{Conclusion}

We presented a full-stack framework for evaluating adaptive personalization of educational readings with theory-grounded simulated learners. The framework links ontology-grounded content generation, KC-level \BKT mastery tracking, \Atlas ontology authoring, \CI-based comprehension, \DIME-style reader heterogeneity, \KReC-style misconception revision, and score-based answer selection over explicit learner memory. Across three sampled subject ontologies, adaptive reading yielded reliable gains in computer science, smaller inconclusive gains in chemistry, and neutral to slightly negative observed outcomes in biology.

\section*{Limitations}

\rev{The framework has several limitations. First, simulated learners are not substitutes for real students. The simulator is designed for inspectable offline policy screening, not as a validated estimator of classroom effect sizes, and it has not yet been calibrated against human response distributions, learning curves, distractor choices, misconception persistence, or subgroup heterogeneity.}

\rev{Second, the modeled instructional setting is intentionally narrow: learners read independently, assessments provide observations but no direct instruction, and peer interaction, instructor intervention, classroom discussion, feedback, and retrieval-practice effects are outside the simulation. This isolates reading as the intervention, but it does not approximate courses where instructors diagnose errors, peers explain concepts, or quizzes teach through feedback and retrieval practice.}

\rev{Third, the ontology and labeling pipeline are central sources of uncertainty. The ontologies are corpus-grounded and built from the same Wikibooks textbooks used for retrieval and generation. Errors in LO/KC granularity, chunk labels, retrieved evidence, or item alignment can propagate through generation, adaptation, and evaluation, so cross-domain differences may reflect content-pipeline quality as well as the adaptive policy. Porting to a new course therefore still requires domain review of the ontology, labels, generated readings, and assessment items rather than only rerunning the automatic pipeline.}

\rev{Fourth, the simulator uses simplified operational heuristics: clarity is approximated from length, refutation strength from lexical cues, and the encoding and response equations use fixed coefficients. Sensitivity to coefficient perturbations, alternative detectors, and ablations remains untested.}

\rev{Fifth, the readability signal is deliberately shallow. New Dale--Chall provides an open, reproducible proxy based on word familiarity and sentence length \citep{chall1995readability}. It was used because the learner model includes vocabulary and background-knowledge factors: Flesch Reading Ease is driven by average sentence length and average syllables per word \citep{flesch1948new}, whereas New Dale--Chall incorporates word familiarity in addition to sentence length. This makes New Dale--Chall a closer operational proxy for whether a passage contains vocabulary likely to be unfamiliar to a reader, including short domain terms that syllable-based formulas may not penalize. However, it does not model discourse cohesion, diagrams, mathematical notation, domain-specific prior vocabulary, or whether a learner already knows terms such as \emph{ATP}, \emph{loop}, or \emph{ion}. Future work should compare New Dale--Chall with Flesch-Kincaid, domain-vocabulary measures, and learned reader--text difficulty models.}

\rev{Sixth, score-based answer selection supports bounded competence and auditability, but it may under-represent open-ended reasoning, explanation quality, and strategic test-taking. We also do not compare against constrained LLM-based student simulators or other cognitive simulators, and preprocessing artifacts may depend on the embedding and instruction models used.}

\rev{Finally, the saved result format preserves learner-level paired trajectories for accuracy and \BKT gain, but only condition-level summaries for hidden mastery and misconception endpoints. Future runs should preserve learner-level hidden-state trajectories for paired analysis.}

\bibliography{custom}

\end{document}